\documentclass[a4paper,10pt,oneside]{article}
\usepackage [english]{babel}
\usepackage [autostyle, english = american]{csquotes}
\MakeOuterQuote{"}
\usepackage[margin=0.7in]{geometry}
\usepackage[utf8]{inputenc}
\usepackage{indentfirst} % Красная строка
\usepackage{url}
\usepackage{cite}
\usepackage{graphicx}
\RequirePackage{caption}
\DeclareCaptionLabelSeparator{defffis}{. }
\usepackage{array}
\usepackage{tabularx}
\providecommand{\keywords}[1]
{
  \small    
  \textbf{Keywords:} #1
}
\usepackage{microtype}
\usepackage{hyperref}
\title{
	\Large \textbf{Distributional semantic modeling: \\
	a revised technique to train term/word vector space models applying the ontology-related approach}
}
\date{\vspace{-5ex}}
\author{
	\begin{tabular}{c}
		Oleksandr Palagin \\ [0pt]
			\small Academician of NAS of Ukraine \\ [-2pt]
			\small \href{mailto:palagin.o.v@nas.gov.ua}{palagin.o.v@nas.gov.ua} \\ [-3pt]
			\tiny \url{https://orcid.org/0000-0003-3223-1391} \\ \\
		Kyrylo Malakhov \\ [0pt]
			\small Junior Researcher \\ [-2pt]
			\small \fbox{\textbf{\href{mailto:malakhovks@nas.gov.ua}{malakhovks@nas.gov.ua}}} \\ [-3pt]
			\tiny \url{https://orcid.org/0000-0003-3223-9844}
	\end{tabular} \and
	\begin{tabular}{c}
		Vitalii Velychko \\ [0pt]
			\small PhD, Senior Researcher \\ [-2pt]
			\small \href{mailto:aduisukr@gmail.com}{aduisukr@gmail.com} \\ [-3pt]
			\tiny \url{https://orcid.org/0000-0002-7155-9202} \\ \\
		Oleksandr Shchurov \\ [0pt] 
			\small Software Engineer \\ [-2pt]
			\small \href{mailto:alexlug89@gmail.com}{alexlug89@gmail.com} \\ [-3pt]
			\tiny \url{https://orcid.org/0000-0002-0449-1295} \\
	\end{tabular} \and 
	\scriptsize	
	V.M. Glushkov Institute of Cybernetics \\[-5pt]
	\scriptsize
	The National Academy of Sciences of Ukraine \\[-5pt]
	\scriptsize	
	Microprocessor Technology Department. 40 Glushkov ave., Kyiv, Ukraine, 03187\\[-5pt]
	\scriptsize \href{tel:380445263348}{+38 044 526 33 48}
}
\begin{document}
\maketitle
\begin{abstract}
We design a new technique for the distributional semantic modeling with a neural network-based approach to learn distributed term representations (or term embeddings) -- term vector space models as a result, inspired by the recent ontology-related approach (using different types of contextual knowledge such as syntactic knowledge, terminological knowledge, semantic knowledge, etc.) to the identification of terms (term extraction) and relations between them (relation extraction) called \emph{semantic pre-processing technology} -- \emph{SPT}. Our method relies on automatic term extraction from the natural language texts and subsequent formation of the problem-oriented or application-oriented (also deeply annotated) text corpora where the fundamental entity is the term (includes non-compositional and compositional terms). This gives us an opportunity to changeover from distributed word representations (or word embeddings) \emph{to distributed term representations (or term embeddings)}. This transition will allow to generate more accurate semantic maps of different subject domains (also, of relations between input terms -- it is useful to explore clusters and oppositions, or to test your hypotheses about them). The semantic map can be represented as a graph using \emph{Vec2graph} -- a Python library for visualizing word embeddings (term embeddings in our case) as dynamic and interactive graphs. The Vec2graph library coupled with term embeddings will not only improve accuracy in solving standard NLP tasks, but also update the conventional concept of automated ontology development. The main practical result of our work is the development kit (set of toolkits represented as web service APIs and web application), which provides all necessary routines for the basic linguistic pre-processing and the semantic pre-processing of the natural language texts in Ukrainian for future training of term vector space models.\end{abstract}
\section*{Introduction}
\addcontentsline{toc}{section}{Introduction}
	Distributional semantic modeling (word embeddings) are now arguably the most popular way to computationally handle lexical semantics. The identification of terms (non-compositional and compositional) that are relevant to the domain is a vital first step in both the automated ontology development and natural language processing tasks. This task is known as term extraction. For ontology generation, terms are first found and then relations between them are extracted. In general, a term can be said to refer to a specific concept which is characteristic of a domain or sublanguage. We propose a new technique for the distributional semantic modeling applying the ontology-related approach. This technique will give us an opportunity to changeover from distributed word representations to distributed term representations. This transition will allow to generate more accurate semantic maps of different subject domains (also, of relations between input terms -- it is useful to explore clusters and oppositions, or to test your hypotheses about them).
\section*{Background. The distributed numerical feature representations of words (word embeddings): word2vec, fastText, ELMo, Gensim}
\addcontentsline{toc}{section}{Background. The distributed numerical feature representations of words (word embeddings): word2vec, fastText, ELMo, Gensim}
	The distributed numerical feature representations of words (word embeddings) and word vector space models, as a result, are well established in the field of computational linguistics and have been here for decades (see \cite{TurneyPeter2010, ganegedara2018natural} for an extensive review). However, recently they received substantially growing attention. Learning word representations lies at the very foundation of many natural language processing (NLP) tasks because many NLP tasks rely on good feature representations for words that preserve their semantics as well as their context in a language. For example, the feature representation of the word \emph{car} should be very different from \emph{fox} as these words are rarely used in similar contexts, whereas the representations of \emph{car} and \emph{vehicle} should be very similar. In distributional semantics, words are usually represented as vectors in a multi-dimensional space of their contexts. Semantic similarity between two words is then calculated as a cosine similarity between their corresponding vectors; it takes values between -1 and 1 (usually only values above 0 are used in practical tasks). 0 value roughly means the words lack similar contexts, and thus their meanings are unrelated to each other. 1 value means that the words' contexts are identical, and thus their meaning is very similar. Word vector space models have been applied successfully to the following tasks: finding semantic similarity between words and multi-word expressions \cite{kutuzov2015textsin}; word clustering based on semantic similarity \cite{kutuzov2014semantic}; automatic creation of thesauri and bilingual dictionaries; lexical ambiguity resolution; expanding search requests using synonyms and associations; defining the topic of a document; document clustering for information retrieval \cite{kutuzov2014semantic}; data mining and named entities recognition \cite{katharina-siencnik-2015-adapting}; creating semantic maps of different subject domains and word embeddings graphs \cite{zhordaniya2020vec2graph}; paraphrasing; sentiment analysis \cite{maas2011learning}. Despite of fundamental differences in ontology-related and neural network-based approaches, vector space models can be used as a part of ontology engineering methodologies \cite{palagin2011technique, palagin2012ontolohichni} as well as a part of ontology engineering development kits \cite{palagin2012ontolohichni, palagin2014development, velychko2014integrated, palagin2012toproblem}.
	\par
		Arguably the most important applications and tools of machine learning in text analysis now are \emph{word2vec} \cite{mikolov2015computing, word2vec} and \emph{fastText} \cite{fasttext} with its Continuous Skip-Gram (CSG) and Continuous Bag of Words (CBOW) algorithms proposed in \cite{mikolov2013efficient, bojanowski2017enriching, joulin2016fasttext}, which allow fast training on huge amounts of raw or pre-processed linguistic data.
	\par
		"You shall know a word by the company it keeps" -- this statement, uttered by J.R. Firth \cite{britannica1957encyclopaedia} in 1957, lies at the very foundation of word2vec, as word2vec techniques use the context of a given word to learn its semantics. Word2vec is a groundbreaking approach that allows to learn the meaning of words without any human intervention. Also, word2vec learns numerical representations of words by looking at the words surrounding a given word. The magic of word2vec is in how it manages to capture the semantic representation of words in a vector. The papers, Efficient Estimation of Word Representations in Vector Space \cite{mikolov2013efficient}, Distributed Representations of Words and Phrases and their Compositionality \cite{mikolov2013distributed}, and Linguistic Regularities in Continuous Space Word Representations \cite{mikolov-etal-2013-linguistic} lay the foundations for word2vec and describe their uses. There are two main algorithms to perform word2vec training, which are the CBOW and CSG models. The underlying architecture of these models is described in the original research paper, but both of these methods involve in understanding the context which we talked about before. The papers written by Mikolov and others provide further details of the training process, and since the code is public, it means we know what’s going on under the hood. 
	\par 
		FastText \cite{fasttext, bojanowski2017enriching, mikolov2013distributed} is a library for efficient learning of word representations and sentence classification. It is written in C++ and supports multiprocessing during training. FastText allows to train supervised and unsupervised representations of words and sentences. These representations (embeddings) can be used for numerous applications from data compression, as features into additional models, for candidate selection, or as initializers for transfer learning. FastText supports training CBOW or CSG models using negative sampling, softmax or hierarchical softmax loss functions. The main difference from word2vec that fastText can achieve really good performance for word representations and sentence classification, especially in the case of rare words by making use of character-level information \cite{bojanowski2017enriching, joulin2016fasttext, joulin2016bag}. Each word is represented as a bag of character \(n\)-grams in addition to the word itself, so for example, for the word \emph{matter}, with \(n=3\), the fastText representations for the character \(n\)-grams are \(<ma, mat, att, tte, ter, er>\). \(<and>\) are added as boundary symbols to distinguish the \(n\)-gram of a word from a word itself, so for example, if the word \emph{mat} is part of the vocabulary, it is represented as \(<mat>\). This helps preserve the meaning of shorter words that may show up as \(n\)-grams of other words. Inherently, this also allows you to capture meaning for suffixes/prefixes.
	\par
		The length of \(n\)-grams you use can be controlled by the \(-minn\) and \(-maxn\) flags for minimum and maximum number of characters to use respectively. These control the range of values to get \(n\)-grams. The model is considered to be a bag of words model because aside of the sliding window of n-gram selection, there is no internal structure of a word that is taken into account for featurization, i.e. as long as the characters fall under the window, the order of the character \(n\)-grams does not matter. You can also turn \(n\)-gram embeddings completely off as well by setting them both to 0. This can be useful when the ‘words’ in your model aren’t words for a \(n\)-grams would not make sense. The most common use case is when you’re putting in ids as your words. During the model update, fastText learns weights for each of the \(n\)-grams as well as the entire word token.
	\par
		The context is important. One of the biggest breakthroughs in distributional semantic modeling is the Embeddings from Language Models (ELMo) representations \cite{peters2018} -- developed in 2018 by AllenNLP \cite{allenNLP}, it goes beyond traditional embedding techniques. It uses a deep, bi-directional LSTM model to create contextualized word representations. Rather than a dictionary of words and their corresponding vectors, ELMo analyses words within the context that they are used. It is also character-based, allowing the model to form representations of out-of-vocabulary words. This, therefore, means that the way ELMo is used is quite different from word2vec or fastText. Rather than having a dictionary ‘look-up’ of words and their corresponding vectors, ELMo instead creates vectors on-the-fly bypassing text through the deep learning model.
	\par
		While the original C code \cite{word2vec} implementation of word2vec released by Google and Mikolov does an impressive job, recently, there is “state of the art” open-source Python library Gensim \cite{gensim} for unsupervised topic modeling and natural language processing, using more efficient implementations of modern statistical machine learning algorithms \cite{luo2015learning, luo2014study, mnih2013learning}. Gensim includes streamed parallelized implementations of fastText, word2vec, and doc2vec algorithms. The primary features of Gensim are its memory-independent nature, multicore implementations of latent semantic analysis, latent Dirichlet allocation, random projections, hierarchical Dirichlet process HDP, as well as the ability to use latent semantic analysis LSA and latent Dirichlet allocation LDA on a cluster of computers. It also seamlessly plugs into the Python scientific computing ecosystem and can be extended with other vector space algorithms.
\section*{\raggedright The conventional technique to train word vector space models}
\addcontentsline{toc}{section}{The conventional technique to train word vector space models}
	\par 
		The review of the most common papers and books in the distributional semantics research area (in particular \cite{ganegedara2018natural, srinivasa2018natural, kadurin2018, goyal2018deep, maynard2016natural}) allowed to create a typical technique for the distributional semantic modeling with a neural network-based approach to learn word embeddings (word vector space models of semantics as a result) represented in figure 1. Let’s briefly review it.
	\begin{figure}[ht]
		\centering
  		\includegraphics[width=\textwidth]{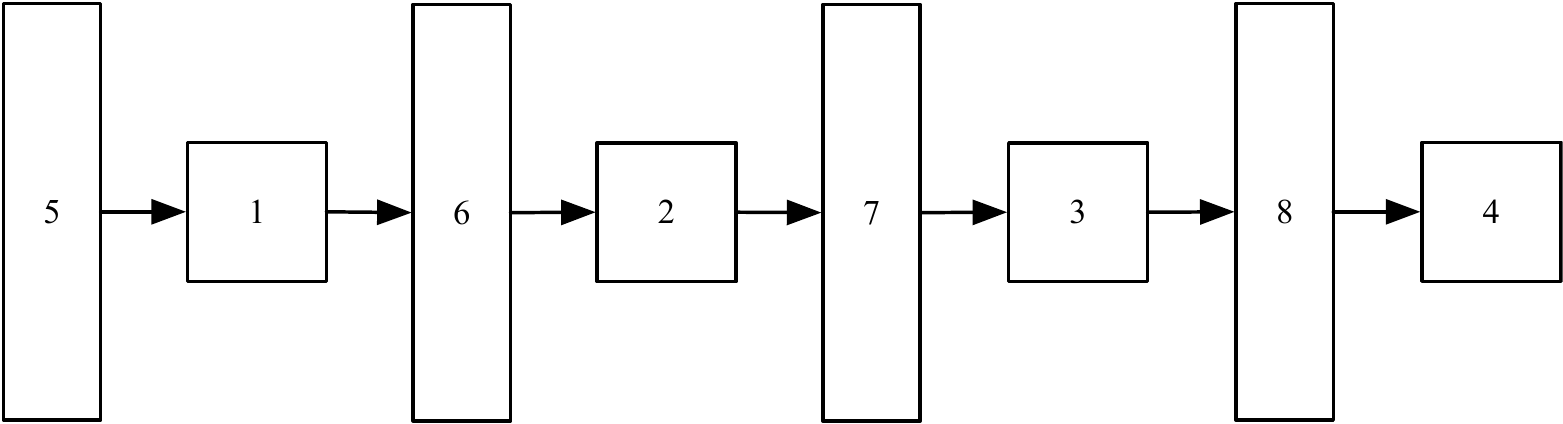}
  		\caption{A typical technique for the distributional semantic modeling with a neural network-based approach to learn word embeddings (word vector space models of semantics as a result).}
	\end{figure}
	\par
		On the practice, a typical technique for the distributional semantic modeling with a neural network-based approach to learn word embeddings (word vector space models of semantics as a result) consists of next components (includes technologies, pipelines and data entities/sources).
	\begin{enumerate}
			\item \emph{Dataset (not annotated text corpus) construction/creation technology.}
				\par
					The secret to getting word2vec, fastText or ELMo algorithms really working for you is to have lots and lots of text data in the relevant domain. For example, if your goal is to build a sentiment lexicon, then using a dataset from the medical domain or even Wikipedia may not be effective. Nevertheless, the most common way to construct the universal text corpus is to use a publicly available and sufficiently large Wikipedia (or other MediaWiki-based) database dump consists of Wikipedia article texts (also you can choose the language of the articles depending on your needs). The original Wikipedia dump \cite{wikimedia} that can be downloaded is in XML format and the structure is quite complex. Thus we need to use an extractor tool to parse it. Python package Gensim provides easy to use application programming interface (API) \emph{genism.corpora.wikicorpus} \cite{gensim} for that purpose and uses multiprocessing internally to parallelize the work and process the dump more quickly. Dataset (not annotated text corpus) represented in the simple text format is the result of applying this technology.
			\item \emph{NLP pre-processing pipeline.}
				\par
					An NLP pre-processing pipeline of Dataset (not annotated text corpus) typically consists of the following components \cite{maynard2016natural}.
				\begin{itemize}
					\item \emph{Tokenization} -- is the task of splitting the input text into very simple units, called tokens, which generally correspond to words, numbers and symbols, and are typically separated by white space in English for example, Tokenization is a required step in almost any linguistic processing application, since more complex algorithms such as part of speech taggers mostly require tokens as their input, rather than using the raw text. Consequently, it is important to use a high-quality tokenizer, as errors are likely to affect the results of all NLP components in the pipeline.
					\item \emph{Sentence splitting} -- or sentence detection is the task of separating the text into its constituent sentences. This typically involves determining whether punctuation (full stops, commas, exclamation marks, and question marks) denote the end of a sentence or something else (quoted speech, abbreviations, etc.).
					\item \emph{Part of speech (POS) tagging} -- is concern with tagging words with their parts of speech (e.g., noun, verb, adjective). These basic linguistic categories are typically divided into quite fine-grained tags, distinguishing for instance between singular and plural nouns, and different tenses of verbs. For languages other than English, for example, Ukrainian, gender may also be included in the tag.
					\item \emph{Elements of the morphological analysis (stemming, lemmatization, filtering out common stopwords)} -- essentially concern the identification and classification of the linguistic units of a word, typically breaking the word down into its root form and an affix. Lemmatization will reduce vocabulary (for word2vec, fastText, ELMo) and increase text coherence.
					\item \emph{Elements of the syntactic parsing (including chunking).} Syntactic parsing is concerned with analyzing sentences to derive their syntactic structure according to a grammar. Essentially, parsing explains how different elements in a sentence are related to each other, such as how the subject and object of a verb are connected. Also base chunking (or shallow parsing) is required to get noun phrases and verb phrases for a simple compositional vocabulary (for word2vec, fastText, ELMo) creation. Note that there is a \emph{gensim.models.phrases module} of the Python Gensim package, which lets you automatically detect phrases longer than one word. Using phrases, you can learn a model where words (entities) are actually multiword expressions, such as \emph{“the-national-academy”} or \emph{“financial-crisis”}. An annotated text corpus is the result of applying this pipeline.
				\end{itemize}
				\par 
					The most common way to implement NLP pre-processing pipeline is to use open-source linguistic pre-processing toolkits such as spaCy \cite{spacy}, NLTK \cite{nltk}, StanfordNLP \cite{stanfordNLP}. spaCy is a free, open-source library for advanced NLP) in Python. It can be used to build information extraction or natural language understanding systems or to pre-process text for deep learning and computational linguistics purposes. Following are the features of spaCy (the Ukrainian language is not supported for now): Non-destructive tokenization; support for more than 21 natural languages; 6 statistical models for 5 languages; Pre-trained word vectors (word embeddings); POS tagging; named entity recognition (NER); labeled dependency parsing; syntax-driven sentence segmentation; built-in visualizers for syntax and NER; export to NumPy data arrays; efficient binary serialization; robust, rigorously evaluated accuracy, etc. spaCy is designed specifically for production use and helps you build applications that process and “understand” large volumes of text. On the other hand, NLTK’ and StanfordNLP’ primary focus is to give students and researchers a toolkit to play around with computational linguistics algorithms is not a production environment.
			\item \emph{Word vector space models (word embeddings) training technology.}
				\par
					Gensim open-source Python library is the most suitable to implement the process of training the new word vector space models. Gensim package has the following API (modules) for this purpose \cite{gensimAPI}: \emph{genism.models.word2vec} (this module implements the word2vec family of algorithms, using highly optimized C routines, data streaming, and pythonic interfaces; the word2vec algorithms include CSG and CBOW models, using either hierarchical softmax or negative sampling); \emph{genism.models.fasttext} (this module allows training word embeddings from a training corpus with the additional ability to obtain word vectors for out-of-vocabulary words; this module contains a fast native C implementation of fastText with Python interfaces); \emph{genism.models.doc2vec} (this module implements learning paragraph and document embeddings via the distributed memory and distributed bag of words models from \cite{le2014distributed}); \emph{genism.models.keyedvectors} (this module implements word vectors and their similarity look-ups; since trained word vectors are independent from the way they were trained, they can be represented by a standalone structure, as implemented in this module). Another important thing about using CSG and CBOW algorithms are hyperparameters values. The detailed review and recommendations of hyperparameters represented in \cite{caselles2018word2vec, rong2014word2vec}.
			\item \emph{Word vector space model (word embeddings) processing (load/serialization and usage) technology.}
				\par
					Deploying deep learning models in the production environment is challenging. There are different ways you can have a model deployed: loading model directly into application (this option essentially considers the model a part of the overall application and hence loads it within the application); calling an API (this option involves making an API and calling the API from your application, this can be done in several different ways, for example, via the Kubernetes open-source container orchestration system for automating application deployment); Serverless cloud-computing execution model (the cloud provider runs the server, and dynamically manages the allocation of machine resources, such as AWS Lambda function as a service solution \cite{awsML}); custom representational state transfer REST API with Flask/Django Python packages from scratch (this option could possibly be combined with Docker and firefly \cite{fireflyDocs} Python package as well).
			\item \emph{Sources of text documents and corpora.}
				\par
					These are analog texts, the Internet, corpora of Wikipedia texts, electronic collections of text documents, databases, etc.
			\item \emph{Dataset (not annotated text corpus) entity.}
				\par
					The most common datasets include an entire corpus of Wikipedia texts, the common crawl dataset \cite{commonCrawl}, or the Google News Dataset or use the Dataset Search \cite{googleDatasetSearch} toolkit from Google. Note if your future application is specific to a certain domain the dataset must be relevant.
			\item \emph{Annotated text corpus entity.}
				\par
					Annotation consists of the application of a scheme to texts. Annotations may include structural markup, POS tagging, parsing, and numerous other representations.
			\item \emph{Word vector space model (word embeddings or distributional semantic model) entity.}
	\end{enumerate}
\section*{\raggedright The revised technique to train term embeddings (term vector space models as a result) applying the ontology-related approach}
\addcontentsline{toc}{section}{The revised technique to train term embeddings (term vector space models as a result) applying the ontology-related approach}
	\par
		In this section, we propose a new technique for the distributional semantic modeling applying the ontology-related approach. Our method relies on automatic term extraction from the natural language texts and subsequent formation of the problem-oriented or application-oriented (also deeply annotated) text corpora where the fundamental entity is the term (includes non-compositional and compositional terms). This technique will give us an opportunity to changeover from distributed word representations (or word embeddings) to distributed term representations (or term embeddings). This transition will allow generating more accurate \emph{semantic maps} of different subject domains (also, of relations between input terms -- it is useful to explore clusters and oppositions or to test your hypotheses about them). The semantic map can be represented as a graph using, for example, Vec2graph \cite{zhordaniya2020vec2graph, vec2graphGithub} -- a Python library for visualizing word embeddings (term embeddings in our case) as dynamic and interactive graphs. Using the Vec2graph library coupled with term embeddings will not only improve accuracy in solving standard NLP tasks but also update the conventional concept of automated ontology development \cite{maynard2016natural} (which comprises three related components: learning, population and refinement) despite the fundamental differences between the ontology-related approach and distributional semantic modeling techniques (that relies on top of the statistical semantics). Ontology learning (or generation) denotes the task of creating an ontology from scratch and mainly concerns the task of defining the concepts -- \emph{terms} and generating relevant relations between them. The ontology population consists of adding instances to an existing ontology structure (as created by the ontology learning task, for instance). Ontology refinement involves adding, deleting, or changing new terms, relations, and/or instances in an existing ontology. Ontology learning may also be used to denote all three tasks, in particular where the tasks of learning and population are performed via a single methodology. For all three components of ontology development, the starting point is typically a large corpus of unstructured text (which may be the entire web itself, or a set of domain-specific documents). For now, the automated ontology development is beyond the scope of this paper but will be concerned in future researches.
	\par
		Figure 2 shows the new technique to train distributed term representations (or term embeddings) -- term vector space model as a result.
	\begin{figure}[ht]
		\centering
  		\includegraphics[width=\textwidth]{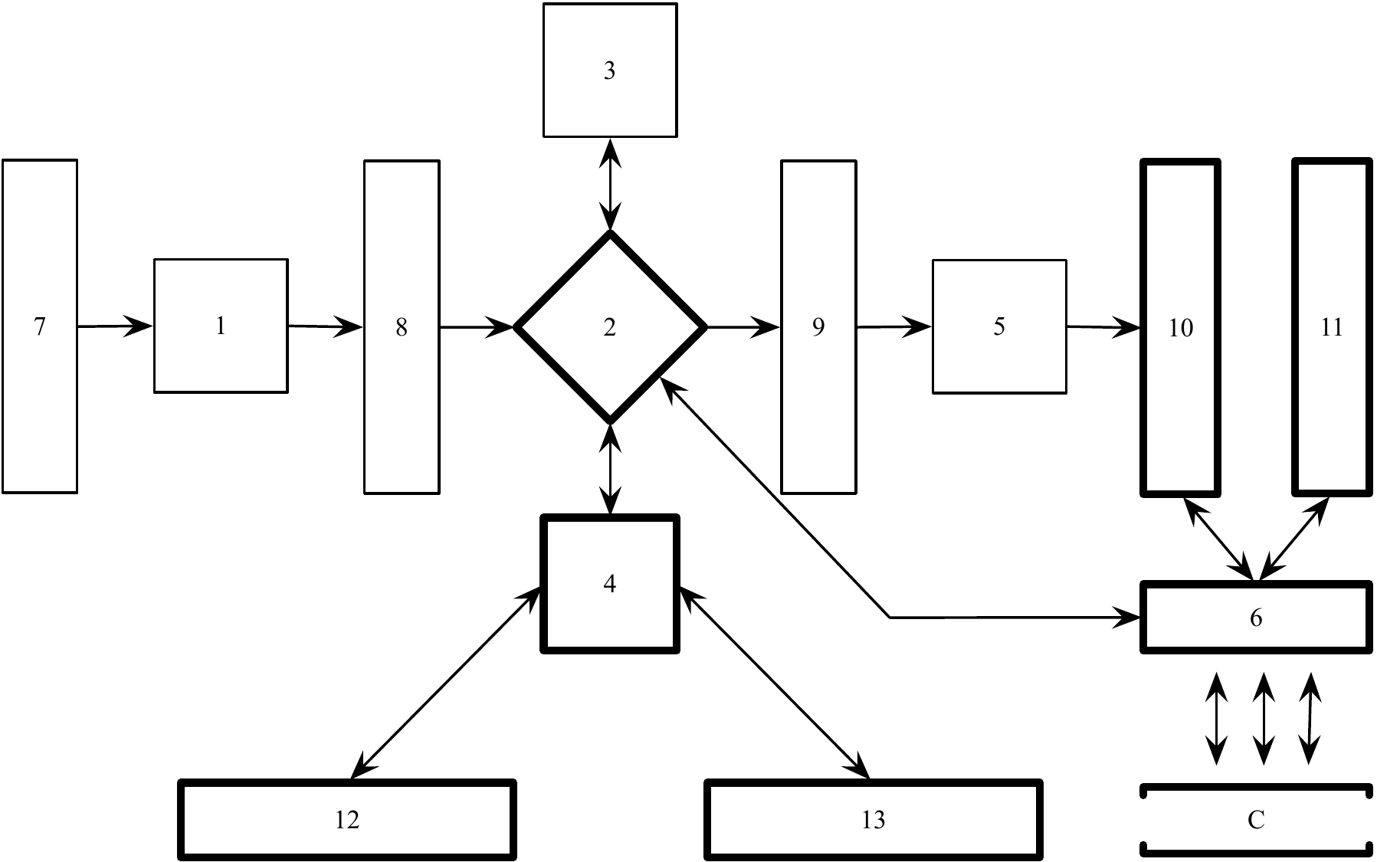}
  		\caption{The new technique to train distributed term representations (or term embeddings) -- term vector space models as a result.}
	\end{figure}
	\par
		The new consists of next components (includes technologies, pipelines and data entities/sources).
	\begin{enumerate}
		\item \emph{Dataset} (not annotated text corpus) construction/creation technology. This component complies with the conventional technique.
		\item The single-page \emph{web application} for managing datasets (not annotated and deeply annotated text corpora) and separate text documents processing technologies.
		\item The basic \emph{NLP pre-processing pipeline}.
		\item \emph {Semantic pre-processing technology SPT}.
			\par
				SPT inspired by the recent ontology-related approach (using different types of contextual knowledge such as syntactic knowledge, terminological knowledge, semantic knowledge, etc.) to the identification of terms (term extraction) and relations between them (relation extraction). For languages with advanced morphology, such as Ukrainian and Russian, inflexions and function words are the primary means of expressing syntax in a sentence. Semantic analysis of a sentence is capable of revealing some errors of syntactic structure. Thus, there is an inverse relationship between semantic and syntactic analysis, so it is advisable to combine these two stages of analysis and execute them together in one analytical unit. SPT technology implemented as a web service (server-side web API interface consisting of several exposed endpoints and using JSON for request–response message system) on top of the functions of the “Konspekt” \cite{palagin2008about} utility for the analysis of the Ukrainian and Russian languages. Let us consider the algorithm of the STP analysis (syntactic and semantic analysis), which is implemented in the “Konspekt” utility. To find a connection between separate words, inflectional means are used to express semantic and syntactic relations. An indicator of the morphological dependency between words is inflection. Segments of phrases that encode the relations between content words, and consist of inflexions and function words are called syntactic determinants \cite{gladun1994}. Since several syntactic relations can correspond to one syntactic determinant, the concept of a correlator \cite{gladun1994} is introduced for the uniqueness of determining the relations between words. Correlators additionally include the grammatical attributes of words inside the phrase.
			\par
				The component of the STP analysis (syntactic and semantic analysis) uses the following input data:
			\begin{itemize}
				\item result of the previous stages of text analysis (grapheme and morphological analysis);
				\item dictionary of stems (contains stems of words and their semantic attributes);
				\item list of all possible inflexions of words;
				\item dictionary of determinants (contains syntactic determinants and lists of correlators for each of them);
				\item dictionary of correlators (each correlator consists of the name of the relation and a list of pairs of semantic attributes of words between which this relation can exist).
			\end{itemize}
			\par
				Let us consider the core stages of the natural language SPT analysis (syntactic and semantic analysis) pipeline.
			\par
				First stage. In each word from the sentence, its stem and inflectional component are defined with the help of the dictionary of stems and the list of inflexions. The classification of words in the sentence is implemented on the grammatical attributes respective stems in the dictionary. Possible ambiguities in the stemming process and determining its grammatical attributes are solved by analyzing the attributes of the words that a next in the sentence, and grammatical attributes of the word stem from the dictionary of stems.
			\par
				Second stage. The syntagmas extraction in a sentence begins with a phrase that defines the core relation (the relation between subject and predicate). In the case when such a phrase cannot be extracted, the sentence is analyzed from left to right from the first content word. For the extracted phrase, a syntactic determinant is formed, which consists of function words and inflectional parts of content words of the phrase. If the generated determinant exists in the dictionary of determinants, a list of correlators is selected for it from the dictionary of determinants. In the correlator dictionary, a correlator is searched from the list of correlators selected in the previous stage, taking into account the grammatical attributes of the stems of words in a possible phrase. The correlator that is found determines the type of syntactic and semantic relations between words. The unambiguity of determining such a relation is ensured by the fact that for a particular determinant, the sets of pairs of grammatical attributes for the correlator from its list do not intersect. The defined content words of the sentence are added to a certain phrase by establishing a relation between the new word and one of the words of the processed part of the sentence. This creates a group of related words. It is important to select a word from the syntagma that will be associated with the following words. It should be either the word with the main relation, or the last word of the syntagma. In the case when it is impossible to determine a relation between the new word and the words of the phrase, a new syntagma is created. At the end of the analysis, all syntagmas are combined into one, which reflects the structure of relations between all the words of the sentence. The impossibility of determining a relation between syntagms in a sentence and their combination testifies either to a complex sentence, parts of which are connected (or implicitly linked) to each other, or to incorrect selection of syntagms. If there are several options for syntagma extraction, a return is made to the step of selecting the list of correlators and another option is selected for determining the syntactic and semantic relations. The best relation option is considered the option with the least number of unrelated syntagms in the sentence.
			\par
				To build the conceptual structure of a natural language text, an automatic term extraction procedure is performed. The terms are considered nouns, abbreviations and noun phrases expressed by the scheme: a matched word + noun. In this model, the noun is the main word, and the matched word is dependent and can be expressed as an adjective or a noun \cite{dobrov2003technology}. If a noun is used as a matched word, then it is in the phrase in the genitive case. Phrases may also include prepositions and compositional conjunctions. The number of words in noun phrases for Russian texts ranges from two to fifteen words and averages three words.
			\par
				An automatic terms extraction (includes compositional terms) procedure uses the results of syntactic and semantic analysis of the text. The terms extraction procedure consists of two main steps \cite{velychko2009avto}. At the first step, there is a direct search in the text of words and phrases -- candidates for terms. As one-word (non-compositional) terms, nouns and abbreviations are chosen. Compositional terms are formed using the types of relations between the words of a sentence defined at the previous stage of the text analysis, by gradually adding words to a one-word (non-compositional) term -- a noun. For terms -- noun phrases, the following basic types of relations between words that are part of the phrases are used: object relation, affiliation (between two nouns), defining relationship (between an adjective and a noun), uniformity words (between two nouns or two adjectives). When the adjective is included in the compositional term, the semantic attributes of the adjective is additionally taken into account. For the compositional term, which includes several nouns, terms of shorter length are automatically extracted and the relation between them is determined. A prerequisite for extracting the term is the correspondence of the relations between the words that are part of it, with certain types of relations.
		\item \emph{Term vector space models (term embeddings) training technology.}
			\par
				This component complies with the conventional technique except that the vector space model is trained on the deeply annotated dataset, the fundamental entity of which is the terms (non-compositional and compositional terms).
		\item \emph{Term vector space models (term embeddings) processing/managing technology as a service.}
			\par
				Term vector space models (term embeddings) processing/managing technology implemented as a server-side web API interface named DS-REST-API. The core functions of the DS-REST-API are: to load pre-trained model (Word2vec, fastText, etc.) and prepare it for inference; to calculate semantic similarity between pairs of terms; to find terms semantically closest to the query term (optionally with POS and frequency filters); to perform analogical inference: to find a term \(X\) which is related to the term \(Y\) in the same way as the term \(A\) is related to the term \(B\); to find the center of term cluster formed by your positive terms.
		\item \emph{Sources of text documents and corpora.}
			\par
				These are analog texts, the Internet, corpora of Wikipedia texts, electronic collections of text documents, databases, etc.
		\item \emph{Datasets} (not annotated text corpora) digital repository.
			\par
				This component complies with the conventional technique.
		\item \emph{Deeply annotated datasets} (text corpora) digital repository.
			\par
				This digital repository stores the deeply annotated datasets (the results of SPT processing over not annotated text corpora).
		\item \emph{Term vector space models} digital repository (contains new pre-trained models).
			\par
				This is an internal digital repository for the new pre-trained word embeddings models.
		\item \emph{Sources of pre-trained distributional semantic models (word embeddings).}
			\par
				These are external digital repositories of the pre-trained word embeddings models.
		\item \emph{Semantic dictionary and lexical WordNet database} (this component may also include distributional thesaurus).
			\par
				This component is used to validate of the semantic relations between terms.
		\item \emph{Wikipedia} online encyclopedia.
			\par
				This component is used to validate terms.
	\end{enumerate}
	\par
		\(C\) -- External clients for web APIs of term vector space models processing/managing technology as a service.
\section*{Conclusion}
\addcontentsline{toc}{section}{Conclusion}
	\par
		We design a new technique for the distributional semantic modeling with a neural network-based approach to learn distributed term representations (or term embeddings) -- term vector space models as a result, inspired by the recent ontology-related approach (using different types of contextual knowledge such as syntactic knowledge, terminological knowledge, semantic knowledge, etc.) to the identification of terms (term extraction) and relations between them (relation extraction) called semantic pre-processing technology -- SPT. Our method relies on automatic term extraction from the natural language texts and subsequent formation of the problem-oriented or application-oriented (also deeply annotated) text corpora where the fundamental entity is the term (includes non-compositional and compositional terms). This gives us an opportunity to changeover from distributed word representations (or word embeddings) to distributed term representations (or term embeddings). This transition will allow to generate more accurate semantic maps of different subject domains (also, of relations between input terms -- it is useful to explore clusters and oppositions, or to test your hypotheses about them). The semantic map can be represented as a graph using Vec2graph \cite{vec2graphGithub} -- a Python library for visualizing word embeddings (term embeddings in our case) as dynamic and interactive graphs. The Vec2graph library coupled with term embeddings will not only improve accuracy in solving standard NLP tasks, but also update the conventional concept of automated ontology development. For example, we want to form a graph of the text document (the fundamental entity in this graph will be a non-compositional or compositional term). This text document may be included in the dataset on which the vector space model was trained or not included in it but it must be relevant to the domain of this model. After the SPT processing of the input text document the set of terms is extracted. Using the Vec2graph library \cite{vec2graphGithub} we visualize (in current vector space model) only terms which present in the input text document. The next step will be marking the relations between those terms partially based on the SPT processing results with future manual corrections by the knowledge engineer. Also, this gives us an opportunity to use pre-trained models which not based on term entities. Using pre-trained models (including contextualized ones) avoids the resource-intensive training process.
	\par
		The main practical result of our work is the development kit (set of toolkits represented as web service APIs and web application), which provides all necessary routines for the basic linguistic pre-processing and the semantic pre-processing of the natural language texts in Ukrainian for future training of term vector space models.
	\par
		In \cite{palagin2018pp} we proposed the new class of Current Research Information Systems and related intelligent information technologies. This class supports the main stages of the scientific research and development lifecycle, starting with the semantic analysis of the information of arbitrary domain area and ending with the formation of constructive features of innovative proposals. It was called -- Research and Development Workstation Environment -- the comprehensive problem-oriented information systems for scientific research and development support. As part of the research and development work of the Glushkov Institute of Cybernetics of National Academy of Sciences of Ukraine (Department of Microprocessor Technology) has developed and implemented the software system in its class. It was called -- Personal Research Information System (PRIS) \cite{palagin2018pp} -- the RDWE class system for supporting research in the field of ontology engineering (the automated building of applied ontology in an arbitrary domain area as a main feature), scientific and technical creativity. The set of toolkits represented as web service APIs and web application are integrated in PRIS atomic web services ecosystem.
\bibliographystyle{unsrt}
\bibliography{reference}
\end{document}